\documentclass[runningheads,a4paper]{llncs}

\usepackage{amssymb}
\usepackage{graphicx}

\usepackage{url}
\usepackage{graphicx}
\usepackage{amssymb}
\usepackage{amsmath}
\usepackage[utf8]{inputenc}
\usepackage{multirow}

\newcommand{\keywords}[1]{\par\addvspace\baselineskip
\noindent\keywordname\enspace\ignorespaces#1}
\begin{document}
\mainmatter  
\title{Key Phrase Extraction of Lightly Filtered Broadcast News}

\author{
   Luís Marujo\inst{1,}\inst{2}
   \and Ricardo Ribeiro\inst{2,}\inst{3}
   \and David Martins de Matos\inst{2,}\inst{4}
   \and João P. Neto\inst{2,}\inst{4}
   \and Anatole Gershman\inst{1}
   \and Jaime Carbonell\inst{1}
}


\institute{
   LTI/CMU, USA\\
   \and
   L2F - INESC ID Lisboa, Portugal\\
   \and
   Instituto Universit\'{a}rio de Lisboa (ISCTE-IUL), Portugal\\
   \and
   Instituto Superior T\'{e}cnico - Universidade T\'{e}cnica de Lisboa,
   Portugal\\
   \email{\{luis.marujo, ricardo.ribeiro, david.matos,
     joao.neto\}@inesc-id.pt}\\
   \email{\{anatoleg, jgc\}@cs.cmu.edu}
}
\maketitle

\begin{abstract}
  This paper explores the impact of light filtering on automatic key
  phrase extraction (AKE) applied to Broadcast News (BN). Key phrases
  are words and expressions that best characterize the content of a
  document. Key phrases are often used to index the document or as
  features in further processing. This makes improvements in AKE
  accuracy particularly important. We hypothesized that filtering out
  marginally relevant sentences from a document would improve AKE
  accuracy.  Our experiments confirmed this hypothesis. Elimination of
  as little as 10\% of the document sentences lead to a 2\%
  improvement in AKE precision and recall.  AKE is built over MAUI
  toolkit that follows a supervised learning approach. We trained and
  tested our AKE method on a gold standard made of 8 BN programs
  containing 110 manually annotated news stories. The experiments were
  conducted within a Multimedia Monitoring Solution (MMS) system for
  TV and radio news/programs, running daily, and monitoring 12 TV and
  4 radio channels.

\end{abstract}
\section{Introduction}
\label{intro}

With the overwhelming amount of News video and audio information
broadcasted daily on TV and radio channels, users are constantly
struggling to understand the big picture. Indexing and summarization
provide help, but they are hard for multimedia documents, such as
broadcast news, because they combine several sources of information,
e.g. audio, video, and footnotes. We use light filtering to improve
the indexing, where AKE is a key element.

AKE is a natural language procedure that selects the most relevant
phrases (key phrases) from a text. The key phrases are phrases
consisting of one or more significant words (keywords). They typically
appear verbatim in the text. Light filtering removes irrelevant
sentences, providing a more adequate search space for AKE. AKE is
supposed to represent the main concepts from the text. But even for a
human being, the manual selection of key phrases from a document is
context-dependent and needs to rely more on higher-level concepts than
low-level features. That is why filtering improves AKE.

In general, AKE consists of two steps~\cite{hulth2004,Sarkar2010,Witten1999}: candidate generation and filtering of the phrases selected in the candidate generation phrase. Several AKE methods have been proposed. Most approaches only
use standard information retrieval techniques, such as N-gram
models~\cite{Cohen1995}, word frequency, TFxIDF (term frequency
$\times$ inverse document frequency)~\cite{Salton:1974}, word
co-occurrences~\cite{Matsuo:2004}, PAT tree or suffix-based for
Chinese and other oriental languages~\cite{Chien:1997}. In addition,
some linguistic methods, based on lexical analysis~\cite{Ercan:2007}
and syntactic analysis~\cite{harabagiu:lacatusu:2005}, are used. These
methodologies are classified as unsupervised
methods~\cite{hasan2010conundrums}, because they do not require
training data. On the other hand, supervised methods view this problem
as a binary classification task, where a model is trained on annotated
data to determine whether a given phrase is a key phrase or
not. Because supervised methods perform better, we use them in our
work. In general, the supervised approach uses machine-learning
classifiers in the filtering step (e.g.: C4.5 decision trees~\cite{Medelyan2010},
neural networks~\cite{Sarkar2010}).

All of the above methods suffer from the presence of irrelevant or
marginally relevant content, which leads to irrelevant key phrases. In
this paper, we propose an approach that addresses this problem through
the use of light filtering based on summarization techniques.

A summary is a shorter version of one or more documents that preserves
their essential content. Compression Ratio (CR) is the ratio of the
length of the removed content (in sentences) to the original
length. Light filtering typically involves a CR near 10\%. Light
filtering is a relaxation to the summarization problem because we just
remove the most irrelevant or marginally relevant content. This
relaxation is very important because the summarization problem is
especially difficult when processing spoken documents: problems like
speech recognition errors, disfluencies, and boundaries identification
(both sentence and document) increase the difficulty in determining
the most important information. This problem has been approached using
shallow text summarization techniques such as Latent Semantic Analysis
(LSA)~\cite{gong:liu:2001} and Maximal Marginal Relevance
(MMR)~\cite{carbonell:goldstein:1998}, which seem to achieve
comparable performance to methods using specific speech-related
features~\cite{penn:zhu:2008}, such as acoustic/prosodic features.

This work here addresses the use of light filtering to improve AKE. The
experiments were conducted within a Media Monitoring Solution (MMS)
system. 

This paper is organized as follows: Section 2 presents the overall
architecture; the description of the summarization module included in
the MMS system is the core of Section 3, results are described in
Section 4, and Section 5 draws conclusions and suggests future work.
\section{Overall Architecture}
\label{sect:architecture}

\begin{figure}[bhtp]
  \centering
  \includegraphics[width=\columnwidth]{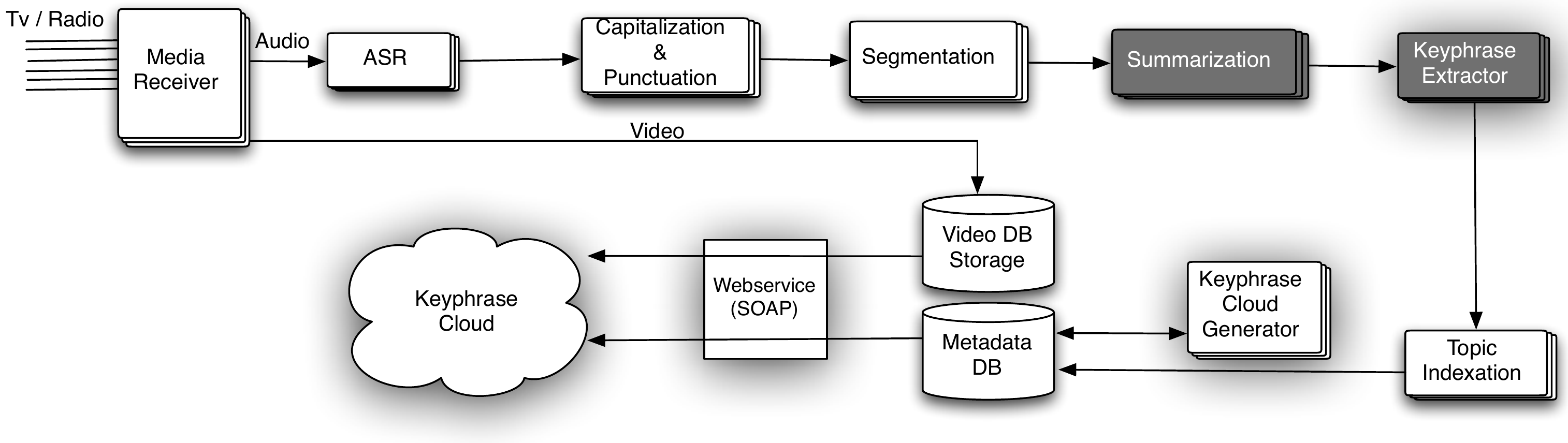}
  \caption{Component view of the system architecture.}
  \label{fig:architecture}
\end{figure}

The main workflow of the complete MMS
system~\cite{Marujo_Interspeech_2011,Neto2011}, depicted in
Fig.~\ref{fig:architecture}, is the following: a Media Receiver
captures and records BN programs from TV and radio. Then, the
transcription is generated and enriched with punctuation and
capitalization. Subsequently, each BN program is automatically
segmented into several stories. News stories are lightly filtered
(90\% of the original size or remains unchanged if the number of
sentences in the summarized version is less than or equal to 3). The
remaining text is passed to the key phrase extraction process. Each
news story is topic-indexed or topic-classified. Finally, each news
story is stored in a metadata database (DB) with the respective
transcription, key phrases, and index, besides program/channel and
timing information. A Key phrase Cloud Generator creates/updates 3D
key phrase cloud based on the interaction with the Metadata DB and
links with the videos that are shown when a user accesses the
system. A 3D key phrase cloud is a tag/word cloud, which is a visual
representation of the most frequently used words in text data. The
most frequent tags are usually displayed in larger fonts in 2D clouds
or at the front in 3D clouds (rotating the 3D cloud allow access to
the less frequent/relevant tags). Typically, tags are keywords or
single words; key phrases extend this concept to several words. 

The gray blocks are the focus of our work.  A summarization
module~\cite{ribeiro:matos:2011}, responsible for the light filtering
step, was included in the workflow and its impact on the key phrase
extraction module is analyzed.
\section{Key Phrase-Cloud Generation Based On Light Filtering}
\label{sect:work}
\subsection{Filtering}
The automatic filtering step applied in this work is performed by a
summarization module that follows a centrality-as-relevance
approach. Centrality-as-relevance methods base the detection of the
most important content on the determination of the most central
passages of the input source(s), considering an adequate input source
representation (e.g.: graph, spatial). Although pioneered in the
context of text summarization, this kind of approaches has drawn some
attention in the context of speech summarization, either by trying to
improve them~\cite{garg:et:al:2009} or using them as
baseline~\cite{lin:yeh:chen:2010}. Even in text summarization, the
number of up-to-date examples is significant.

The summarization model we use~\cite{ribeiro:matos:2011} does not need
training data or additional information. The method consists in
creating, for each passage of the input source, a set containing only
the most semantically related passages, designated support set. Then,
the determination of the most relevant content is achieved by
selecting the passages that occur in the largest number of support
sets. Geometric proximity (Manhattan, Euclidean, Chebyshev are some of
the explored distances) is used to compute semantic
relatedness. Centrality (relevance) is determined by considering the
whole input source (and not only local information), and by taking
into account the presence of noisy content in the information sources
to be summarized. This type of representation diminishes the influence
of the noisy content, improving the effectiveness of the centrality
determination method.
\subsection{Automatic Key Phrase Extraction}

AKE extracts key concepts. Our AKE process was designed to take into
account the extraction of few key phrases (e.g.: 10 used in 3D Key
phrase Cloud) and large number of key phrases (e.g.: 30 used for
indexing). We privileged precision over recall when extracting fewer
key phrases because we want to mitigate visible mistakes in the 3D Key
phrases Cloud. On the other hand, recall gains importance when we
extract many key phrases because we want to have the best coverage
possible. During our experiments, we observed that the most general
and at the same time relevant concepts can be directly linked with an
index topic (examples: soccer/football $\rightarrow$ sports,
PlayStation $\rightarrow$ technology). However, they are frequently
captured by the previous methods with low confidence ($<$50\%). Since
filtering reduces irrelevant content, it increases the confidence of
capturing the best key phrases. The AKE system we
use~\cite{Marujo_Interspeech_2011}, developed for European Portuguese
BN, is an extended version of Maui-indexer toolkit~\cite{Medelyan2010}
(a state-of-art supervised key phrase extraction toolkit), which is in
turn an improved version of KEA~\cite{Witten1999}. Training data is
used to train a machine learning classifier (bagging over C4.5
decision tree). The output is a model that uses extracted features to
classify whether a word or phrase is a key phrase. The same CR
(filtering) is used to train the models and evaluate them at the test
sets. This allows the models to be more robust. The Maui-indexer
feature extraction process was enriched with the following 5 features:
number of characters; the number of named entities using the
MorphoAdorner name recognizer; number of capital letters; count of POS
tags; and probability of the key phrase in a 4-gram domain model
(about 58K unigrams, 700M bigrams, 1.500M trigrams, and 10.000M
4-grams). We have previously demonstrated that these features improved
AKE~\cite{Marujo_Interspeech_2011}.
\section{Evaluation}
\label{sect:evaluation}

\begin{figure}[bhtp!]
  \centering
  \includegraphics[width=.3\columnwidth]{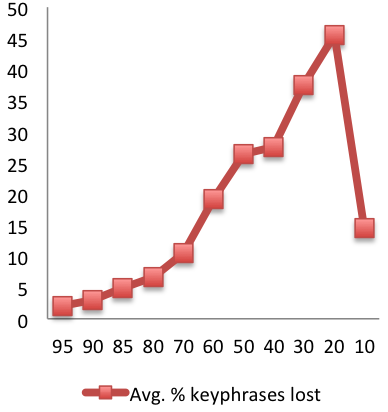}
  \caption{Avg. key phrase percentage lost in
    summarization. The results were obtained when extracting 10
    keyphrases in the test set using 10\% SSC and Manhattan distance.}
  \label{fig:average:loss}
\end{figure}

We used a BN gold standard corpus annotated with the corresponding key
phrases, created in previous work. The gold standard consists of 8 BN
programs transcribed from the European Portuguese ALERT BN
database. The news transcriptions were produced by AUDIMUS, an ASR for
Portuguese, with low WER (14,56\% on average); and punctuated and
capitalized automatically using in-house tools. Those news programs
were automatically split into a total of 110 news stories. Later, each
news story was manually examined to fix segmentation
errors. Afterward, one annotator was asked to extract all key phrases
that represent a relevant concept in each news story. The gold
standard was divided in training (100 news stories containing on
average 24 key phrases and 19 sentences) and test set (10 news stories
containing on average 29 key phrases and 17 sentences).
\begin{figure}[bhtp]
  \centering
  \begin{tabular}{cc}
    \includegraphics[width=.25\columnwidth]{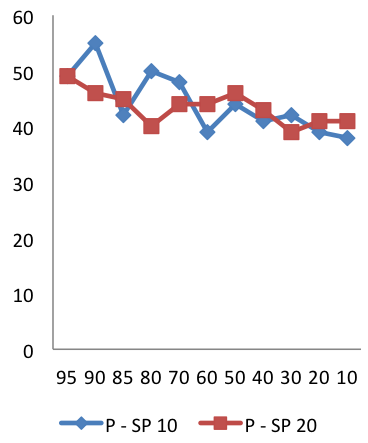} &
    \includegraphics[width=.25\columnwidth]{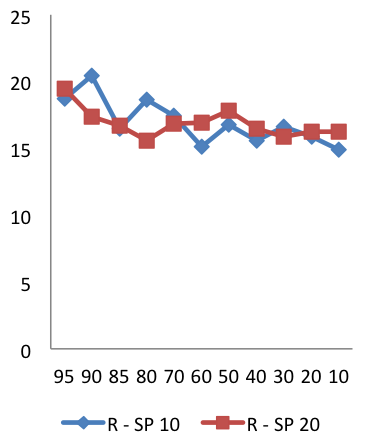}\vspace{-5pt}\\
    \multicolumn{2}{c}{\scriptsize (a)}\\
    \includegraphics[width=.25\columnwidth]{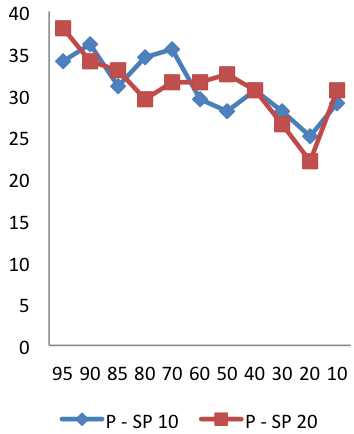} &
    \includegraphics[width=.25\columnwidth]{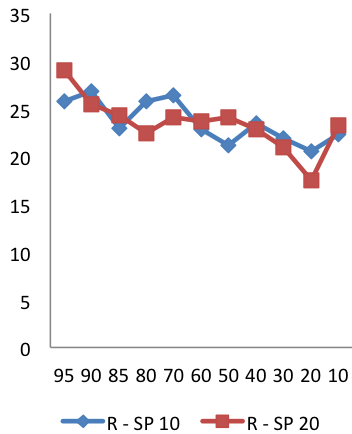}\vspace{-5pt}\\
    \multicolumn{2}{c}{\scriptsize (b)}\\
    \includegraphics[width=.25\columnwidth]{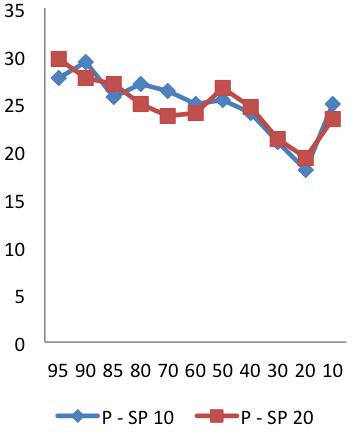} &
    \includegraphics[width=.25\columnwidth]{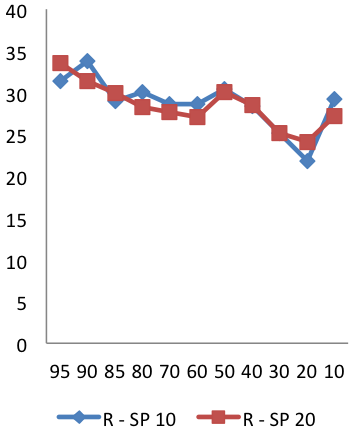}\vspace{-5pt}\\
    \multicolumn{2}{c}{\scriptsize (c)}\\
  \end{tabular}
  \caption{The percentage of the original text in X-axis vs. AKE
    metrics in Y-axis.  The evaluation performed in the test set used
    the Manhattan metric, 10\% and 20\% SSC obtained The Precision and
    Recall extracting: (a) 10, (b) 20 and (c) 30 key phrases.}
  \label{fig:results}
\end{figure}
\begin{table}[h!]
\setlength{\tabcolsep}{2pt}
\centering
\caption{\label{KeyphraseExtractionResults10PerSum} AKE results obtained in the test set using light filtering (p-value $\approx$ 0.1).}
\begin{tabular}{|c|c|c|c|c|c|c|c|}
\hline
\bf{\#Key. Extr.}&\bf{\%orig. text} & \bf{SSC} & \bf{Dist.Metric} &\bf{\#Key.Ident.} & \bf{P} & \bf{R} &\bf{F1}\\
\hline
10 & 100\% & - & - & 5.3 & 53 & 20.63 & 29.7\\\hline
10 & 90\% & 20\% & chebyshev & 4.7 & 47.00 & 18.45 & 26.50\\ \hline
10 & 90\% & 10 & chebyshev & 5.3 & 53.00 & 19.57  & 28.59\\ \hline
\bf{10} & \bf{90\%} & \bf{10\%} & \bf{manhattan} & \bf{5.5} & \bf{55} & \bf{20.45} & \bf{29.81}\\ \hline
10 & 90\% & 5 & manhattan & 5.0 & 45.00 & 17.88 & 26.05\\ \hline
10 & 90\% & 20 & manhattan & 5.3 & 53.00 & 20.67 & 29.71\\ \hline
10 & 90\% & 10\% & minkowski & 4.8 & 48.00 & 18.27 & 26.46\\ \hline
10 & 90\% & 8 & minkowski & 4.6 & 46.00 & 17.45 & 25.30\\ \hline
10 & 90\% & 20 & minkowski & 5.1 & 51.00 & 18.84 & 27.52\\ \hline
10 & 90\% & 20\% & cosine & 5.1 & 51.00 & 19.34 & 28.04\\ \hline
10 & 90\% & 20\% & euclidean & 4.8 & 48 & 18.67 & 26.88\\ \hline
10 & 90\% & 5 & euclidean & 5.0 & 50.00 & 19.37 & 27.93\\ \hline
\hline
\hline
20 & 100\% & - & - & 7.4 & 37 & 28.21 & 32.01 \\\hline
20 & 90\% & 20\% & manhattan & 6.8 & 34 & 25.45 & 29.11\\\hline
20 & 90\% & 20 & manhattan & 5.2 & 52.00 & 19.63 & 28.51\\\hline
20 & 90\% & 10\% & minkowski & 7.1 & 35.5 & 27.74 & 31.14\\\hline
\bf{20} & \bf{90\%} & \bf{8} & \bf{minkowski} & \bf{7.6} & \bf{38.00} & \bf{29.08} & \bf{32.95}\\\hline
20 & 90\% & 20\% & euclidean & 7.5 & 37.5 & 28.43 & 32.34\\ \hline
20 & 90\% & 21 & euclidean & 7.6 & 37.50 & 28.13 & 32.33\\\hline
\hline
\hline
30 & 100\% & - & - & 9.2 & 30.67 & 35.12 & 32.74 \\\hline
30 & 90\% & 10\% & manhattan & 8.8 & 29.33 & 33.75 & 31.39\\\hline 
30 & 90\% & 5 & manhattan & 8.9 & 29.67 & 33.21 & 31.34\\\hline 
30 & 90\% & 20\% & minkowski & 8.6 & 28.67 & 34.48 & 31.31\\\hline 
\bf{30} & \bf{90\%} & \bf{8} & \bf{minkowski} & \bf{9.5} & \bf{31.67} & \bf{35.99} & \bf{33.69}\\\hline 
30 & 90\% & 20\% & euclidean & 8.8 & 29.33 & 32.81 & 30.97\\\hline 
30 & 90\% & 25 & euclidean & 9.2 & 30.67 & 34.57 & 32.50\\\hline 
\hline
\hline
40 & 100\% & - & - & 10.3 & 25.75 & 38.87 & 30.98 \\\hline
40 & 90\% & 10\% & manhattan & 10.1 & 25.25 & 38.44 & 30.48 \\\hline
40 & 90\% & 20\% & manhattan & 9.3 & 23.25 & 35.50 & 28.10 \\\hline
\bf{40} & \bf{90\%} & \bf{8} & \bf{minkowski} & \bf{10.6} & \bf{26.50} & \bf{40.82} & \bf{32.14} \\\hline
40 & 90\% & 20\% & minkowski & 9.6 & 24.00 & 38.00 & 29.42 \\\hline
40 & 90\% & 10\% & euclidean & 9.3 & 23.25 & 35.64 & 28.14 \\\hline
40 & 90\% & 20\% & euclidean & 10.3 & 25.75 & 38.99 & 31.02 \\\hline
\end{tabular}
\end{table}
In our experiments, light filtering improved AKE precision and recall
by 2\%. We have also tested higher CR (Figure \ref{fig:results}) and restricting the summary length to 4 sentences (roughly the average size of a paragraph). However, we did not observed improvements in the results.
The average percentage of key phrases lost by the filtering
process was less than 5\% (Figure \ref{fig:average:loss}).
\section{Conclusions and future work}

This paper explores a novel method to improve key phrase extraction
from BN by using light filtering. The key phrases are extracted to
create a hierarchical 3-layer representation of news. The key phrases
of top news are visualized in tag cloud to allow users to skim their
content and jump to the most relevant news faster.

Based on the results, we show that light filtering improves automatic
key phrase extraction. We included light filtering, constrained to
have at least 4 sentences in the summary in the MMS system. This step
is done before extracting 10 key phrases of each news story. In
addition, we show that even changing the number of key phrases
extracted the light filtering still improves the AKE process. We also
show that filtering up to 50\% of the original size corresponds to
about 26\% in key phrases loss. That corresponds to less than 5\% in
terms of AKE performance metrics degradation. This is an important
result because we take advantage of the summary shown in the MMS
interface to reduce AKE computational resources, such as processing
time, while the AKE performance degradation is very low. Nevertheless,
we create news summaries at both 10\% and 50\% CR to use before the
AKE and to shown in the MMS interface. At the present time, the MMS
interface uses AKE process to identify the 10 top ranked key phrases
from top news from 12 TV and 4 Radio channels and generate the 3D key
phrase cloud. Although 50\% CR seem enough to us, we would like to
analyze in future research what percentages of CR users
prefer. Alternatively, they could prefer to customize this value based
on the amount of time available to interact with the system.

In the future, we plan to augment the centrality-based summarization
with AKE.
\subsection*{Acknowledgments} This work was supported by Carnegie Mellon Portugal Program and under FCT grant SFRH/BD /33769/2009.
\bibliographystyle{splncs03}
\bibliography{paper}

\begin{thebibliography}{10}
\providecommand{\url}[1]{\texttt{#1}}
\providecommand{\urlprefix}{URL }

\bibitem{carbonell:goldstein:1998}
Carbonell, J., Goldstein, J.: {The Use of MMR, Diversity-Based Reranking for
  Reordering Documents and Producing Summaries}. In: ACM SIGIR 1998. pp.
  335--336 (1998)

\bibitem{Chien:1997}
Chien, L.: Pat-tree-based keyword extraction for chinese information retrieval.
  In: ACM SIGIR 1997. pp. 50--58. ACM, New York, NY, USA (1997)

\bibitem{Cohen1995}
Cohen, J.D.: {Highlights : Language- and Domain-Independent Indexing Terms for
  Abstracting Automatic}. English  46(3),  162--174 (1995)

\bibitem{Ercan:2007}
Ercan, G., Cicekli, I.: Using lexical chains for keyword extraction.
  Information Processing \& Management  43(6),  1705 -- 1714 (2007), text
  Summarization

\bibitem{garg:et:al:2009}
Garg, N., Favre, B., Reidhammer, K., Hakkani-T{\"u}r, D.: {ClusterRank: A Graph
  Based Method for Meeting Summarization}. In: {Proceedings of INTERSPEECH
  2009}. pp. 1499--1502. ISCA (2009)

\bibitem{gong:liu:2001}
Gong, Y., Liu, X.: {Generic Text Summarization Using Relevance Measure and
  Latent Semantic Analysis}. In: ACM SIGIR 2001. pp. 19--25. ACM (2001)

\bibitem{harabagiu:lacatusu:2005}
Harabagiu, S., Lacatusu, F.: {Topic Themes for Multi-Document Summarization}.
  In: ACM SIGIR 2005. pp. 202--209. ACM (2005)

\bibitem{hasan2010conundrums}
Hasan, K., Ng, V.: Conundrums in unsupervised keyphrase extraction: making
  sense of the state-of-the-art. In: ACL 2010. pp. 365--373. ACL (2010)

\bibitem{hulth2004}
Hulth, A., Karlgren, J., Jonsson, A., Bostr\"{o}m, H., Asker, L.: Automatic
  keyword extraction using domain knowledge. CICLing pp. 472--482 (2004)

\bibitem{lin:yeh:chen:2010}
Lin, S.H., Yeh, Y.M., Chen, B.: {Extractive Speech Summarization -- From the
  View of Decision Theory}. In: Proceedings of Interspeech 2010. ISCA (2010)

\bibitem{Marujo_Interspeech_2011}
Marujo, L., Viveiros, M., Neto, J.P.: {Keyphrase Cloud Generation of Broadcast
  News}. In: Interspeech 2011. ISCA (September 2011)

\bibitem{Matsuo:2004}
Matsuo, Y., Ishizuka, M.: Keyword extraction from a single document using word
  co-ocurrence statistical information. Inter. Journal on A.I. Tools  13,
  157--170 (2004)

\bibitem{Medelyan2010}
Medelyan, O., Perrone, V., Witten, I.H.: {Subject metadata support powered by
  Maui}. In: Proceedings of JCDL '10. p. 407. ACM Press, New York, USA (2010)

\bibitem{Neto2011}
Neto, J.P., Meinedo, H., Viveiros, M.: A media monitoring solution. In:
  Proceedings of ICASSP 2011. Prague, Czech Republic (2011)

\bibitem{penn:zhu:2008}
Penn, G., Zhu, X.: {A Critical Reassessment of Evaluation Baselines for Speech
  Summarization}. In: Proceeding of ACL-08: HLT. pp. 470--478. ACL (2008)

\bibitem{ribeiro:matos:2011}
Ribeiro, R., de~Matos, D.M.: {Revisiting Centrality-as-Relevance: Support Sets
  and Similarity as Geometric Proximity}. Journal of A.I. Research  42,
  275--308 (2011)

\bibitem{Salton:1974}
Salton, G., Yang, C.S., Yu, C.T.: A theory of term importance in automatic text
  analysis. Tech. rep., Ithaca, NY, USA (1974)

\bibitem{Sarkar2010}
Sarkar, K., Nasipuri, M., Ghose, S.: A new approach to keyphrase extraction
  using neural networks. Inter. Journal of Computer Science Issues  7(2,3),
  16--25 (2010)

\bibitem{Witten1999}
Witten, I., Paynter, G., Frank, E., Gutwin, C., Nevill-Manning, C.: {KEA:
  Practical automatic keyphrase extraction}. In: Proceedings of the fourth ACM
  conference on Digital libraries. pp. 254--255. ACM (1999)

\end{thebibliography}
\end{document}